\documentclass[10pt,twocolumn,letterpaper]{article}

\usepackage{iccv}

\usepackage{amsmath,amsfonts,bm}

\def\eqref#1{equation~\ref{#1}}

\def\1{\bm{1}}

\DeclareMathAlphabet{\mathsfit}{\encodingdefault}{\sfdefault}{m}{sl}
\SetMathAlphabet{\mathsfit}{bold}{\encodingdefault}{\sfdefault}{bx}{n}

\def\gF{{\mathcal{F}}}

\def\gX{{\mathcal{X}}}
\def\gY{{\mathcal{Y}}}
\def\gZ{{\mathcal{Z}}}

\def\shat{\hat{s}}

\def\zhat{\hat{z}}

\DeclareMathOperator*{\argmin}{arg\,min}

\usepackage{times}
\usepackage{epsfig}
\usepackage{graphicx}
\usepackage{amsmath}
\usepackage{amssymb}

\usepackage{algorithm}
\usepackage{algorithmic}

\usepackage{url}
\usepackage{multirow}
\usepackage{diagbox}
\usepackage{tabularx}
\usepackage{caption}
\usepackage{graphicx}
\usepackage{subcaption}
\usepackage{adjustbox}
\usepackage{enumitem}
\usepackage{soul}

\usepackage{xhfill}

\usepackage{subfiles}
\newcommand{\be}{\begin{equation}}
\newcommand{\ee}{\end{equation}}

\usepackage[pagebackref=true,breaklinks=true,letterpaper=true,colorlinks,bookmarks=false]{hyperref}

\iccvfinalcopy 

\ificcvfinal\fi

\begin{document}

\title{Robustness via Cross-Domain Ensembles}

\author{
Teresa Yeo $^{\dagger}$\footnotemark[1] 
\quad\quad
O\u{g}uzhan Fatih Kar $^{\dagger}$\footnotemark[1]
\quad\quad
Alexander Sax $^{\ddagger}$
\quad\quad
Amir Zamir $^{\dagger}$
\vspace{10pt}\\ 
$^\dagger$ Swiss Federal Institute of Technology (EPFL)  \;\;  
$^\ddagger$ UC Berkeley\vspace{10pt}\\ 
\small\url{http://crossdomain-ensembles.epfl.ch/}
}

\maketitle
\ificcvfinal\thispagestyle{empty}\fi

\begin{abstract}
We present a method for making neural network predictions robust to shifts from the training data distribution. The proposed method is based on making predictions via a diverse set of cues (called `middle domains') and ensembling them into one strong prediction. The premise of the idea is that predictions made via different cues respond differently to a distribution shift, hence one should be able to merge them into one robust final prediction. We perform the merging in a straightforward but principled  manner based on the uncertainty associated with each prediction. The evaluations are performed using multiple tasks and datasets (Taskonomy, Replica, ImageNet, CIFAR) under a wide range of adversarial and non-adversarial distribution shifts which demonstrate the proposed method is considerably more robust than its standard learning counterpart, conventional deep ensembles, and several other baselines. 
\end{abstract}

\section{Introduction}

Neural networks deployed in the real world will encounter data with naturally occurring distortions, e.g. motion blur, or adversarial ones. Such changes make up shifts from the training data distribution. While neural networks are able to learn complex functions in-distribution, their predictions are profoundly unreliable under such shifts~\cite{dodge2017study,hendrycks2019benchmarking,szegedy2013intriguing,jo2017measuring}. This presents a core challenge that needs to be solved for these models to be useful in the real world. 

\begin{figure}[!t]
\centering
  \includegraphics[scale=0.100]{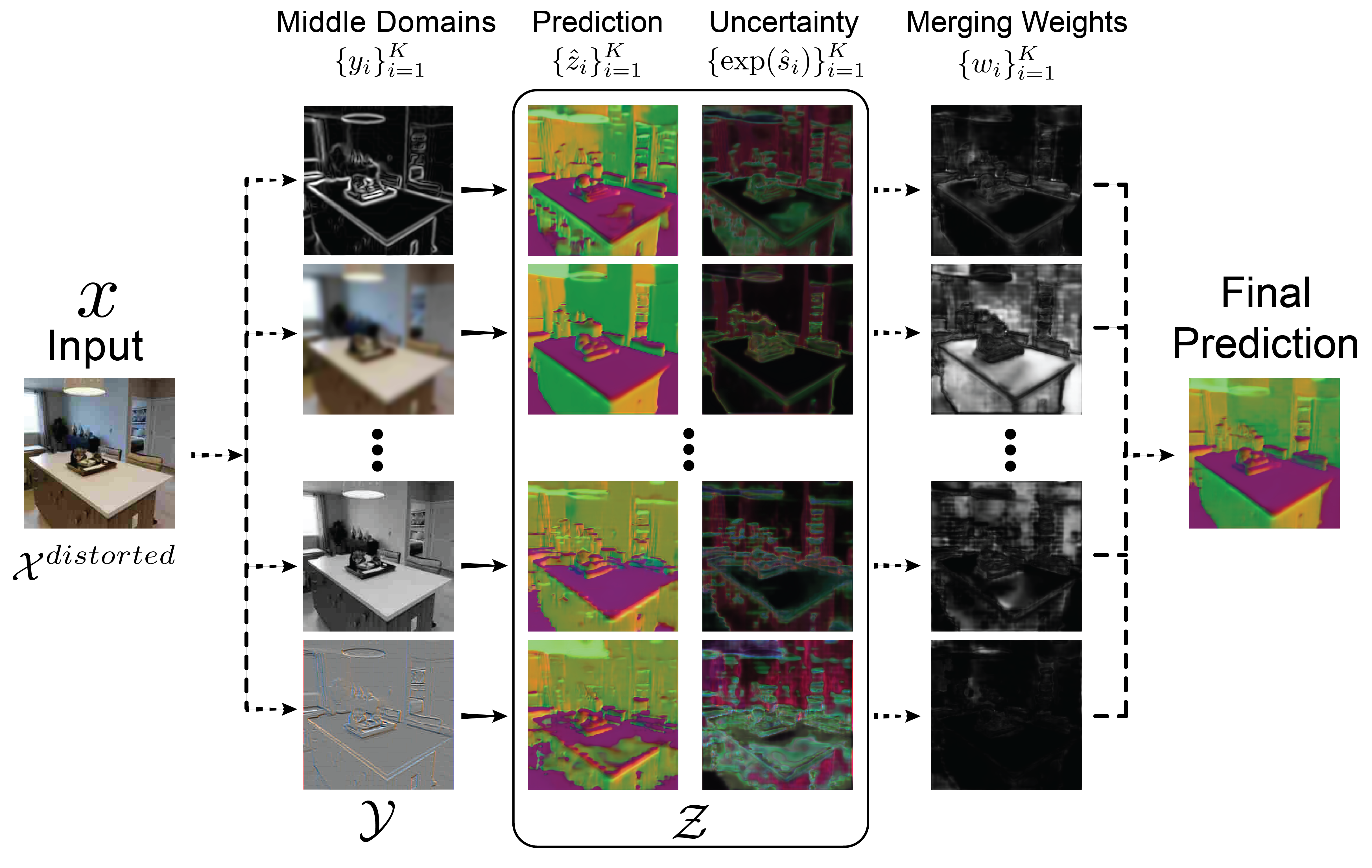} 
\caption{\footnotesize{{\textbf{An overview of the proposed method for creating a robust and diverse ensemble of predictions.} A set of $K$ networks predict a target domain (here surface normals) given an input image that has undergone an unknown distribution shift (here JPEG compression degradation), via $K$ middle domains (e.g. 2D texture edges, low-pass filtering, greyscale image, emboss filtering, etc). The predictions from the $K$ paths are then merged into one final strong prediction using weights that are based on the uncertainty associated with each prediction. This method is shown to be significantly robust against adversarial and non-adversarial distribution shifts for several tasks. In the figure above, solid~(\rule[.5ex]{.5em}{.4pt}) and dashed~(\rule[.5ex]{.25em}{.4pt}~\rule[.5ex]{.25em}{.4pt}) arrows represent learned and analytical functions, respectively.}}}\label{fig:method}
\end{figure}

Suppose we want to learn a mapping from an input domain, e.g. RGB images, to a target domain, e.g. surface normals (see Fig.~\ref{fig:method}). A common approach is to learn this mapping with a \emph{direct} path, i.e. $\textit{RGB} \rightarrow \textit{surface normals}$. Since this path directly operates on the input domain, it is prone to being affected by any slight alterations in the RGB image, e.g. brightness changes. An alternative can be to go through a \emph{middle domain}\footnote{or equivalently ``middle task", as most vision tasks can be viewed as mapping an input onto some other domain.\\
\phantom{....} *~Equal contribution.} that is invariant to that change. For example, the surface normals predicted via the $\textit{RGB} \rightarrow \textit{2D edges} \rightarrow \textit{surface normals}$ path will be resilient to brightness distortions in the input as the 2D edges domain abstracts that away. However, one does not know which middle-domain to use ahead of time as the distortions that a model may encounter are broad and apriori unknown, and some middle domains can be too lossy for certain downstream predictions. These issues can be mitigated by employing an \emph{ensembling} approach where predictions made via a diverse \emph{set} of middle domains are merged into one strong prediction on-the-fly.

This paper presents a general approach for the aforementioned process. We first use a set of $K$ middle domains from which we learn to predict the final domain (Fig.~\ref{fig:method}). Each of the $K$ paths reacts differently to a particular distribution shift due to its inherent biases, so its prediction may or may not degrade severely. Thus, we further estimate the \emph{uncertainty} of each path's prediction which allows us to employ a principled way of combining these predictions into the one final prediction. 

Prior knowledge of the relationship between middle domains is not needed as their contribution to the final prediction is guided by their predicted uncertainties in a fully computational manner independent of the definition of the middle domain. In other words, no manual modification or re-design is needed upon a change in these domains. Moreover, the middle domains we adopt can all be \emph{programmatically extracted}. Thus, this framework does not require any additional supervision/labeling than what a dataset already comes with. The proposed method would be equally applicable if the middle domains were also obtained using a learning based approach, e.g. predicting surface normals from the output of another network such as a depth estimator. We show in Sec.~\ref{sec-exp} that the method performs well insensitive to the choice of middle domains and it generalizes to completely novel non-adversarial and adversarial corruptions.




\section{Related Work}
This work has connections to a number of topics, including ensembling, uncertainty estimation and calibration, inductive bias learning~\cite{baxter2000model}, or works in neuroscience that suggest the brain uses multiple, sometimes partially redundant, cues to perceive~\cite{howard2002seeing,howard1995binocular}. We give an overview of some of them within the constraints of space.

\textbf{Ensembling} allows us to resolve the bias-variance trade-off which states that errors in a models' prediction can be decomposed into bias, variance, and an irreducible data-dependent noise term~\cite{geman1992neural,friedman2001elements}. This is done by combining multiple models with low bias and high variance, e.g. bagging~\cite{dietterich2000ensemble}, or with high bias and low variance, e.g. boosting~\cite{dietterich2000ensemble}, to have predictions with both low bias \textit{and} variance. 


A primary challenge for ensemble methods is to ensure diversity. Sources of diversity include using different initializations~\cite{lakshminarayanan2017simple}, hyperparameters~\cite{wenzel2020hyperparameter} or network architectures~\cite{zaidi2020neural} for the ensemble components, or training the ensemble with additional loss terms~\cite{pang2019improving,kariyappa2019improving,yang2020dverge}. However, under distribution shifts, reduction in performance can stem from an increase in the bias, rather than the variance term~\cite{yang2020rethinking}. Our set of middle domains yields a more diverse ensemble by design and promotes invariance to different distortions to keep bias low~(see Fig.~\ref{fig:method}).

\textbf{Estimating uncertainty:} Uncertainty in a model's prediction can be decomposed into two sources~\cite{der2009aleatory,kendall2017uncertainties}. \emph{Epistemic} uncertainty accounts for uncertainty in the model parameters, while \emph{aleatoric} uncertainty stems from the noise inherent in the data. There are many proposed methods to estimate the former, such as using dropout~\cite{gal2016dropout,srivastava2014dropout}, stochastic variational inference methods~\cite{blundell2015weight,graves2011practical,louizos2017multiplicative,louizos2016structured,wen2018flipout,riquelme2018deep}, ensembling~\cite{lakshminarayanan2017simple}, and consistency energy~\cite{zamir2020robust} where a \emph{single} uncalibrated uncertainty estimate is extracted from consistency of different paths. Most of the existing methods in this area solely estimate uncertainty without using it towards improving the predictions. In contrast, our formulation estimates a calibrated uncertainty for each path \emph{and} uses it to produce a stronger prediction. 


\textbf{Improving calibration with auxiliary datasets:} Neural networks tend to produce outputs that are miscalibrated, i.e. their estimated uncertainty does not reflect the true likelihood of being correct~\cite{guo2017calibration,kuleshov2018accurate}. In particular, their predictions tend to be \emph{overconfident} for unfamiliar examples. This is usually handled by a calibration step. Similar to~\cite{hafner2020noise,hendrycks2018deep,liang2017enhancing,lee2017training}, we use a separate dataset from the one at test time to train the model to output high sigmas (uncertainties) for unfamiliar cases. Previous papers focus on generalizing uncertainties for classification; in Section~\ref{sec:method-noise}, we show this can be extended to dense regression problems.

\textbf{Enforcing consistency constraints} in the context of cross-task predictions involves ensuring that the output predictions remain the same regardless of the intermediate domain~\cite{zamir2020robust,leordeanu2020semi,zhu2017unpaired,xu2018pad}. Particularly in contrast to~\cite{zamir2020robust} which uses (non-probabilistic) training-time consistency constraints to improve a network's prediction and does not have any consolidation mechanism, our goal is to robustify the final prediction by \emph{merging the output of multiple prediction paths at the test time}. Our formulation and the \emph{training-time} consistency constraints are complimentary.

\textbf{Robustness via data augmentation:} One approach to addressing robustness involves using data augmentation during training~\cite{madry2017towards,zhang2017mixup,hendrycks2019augmix,kim2020learning,ashukha2020pitfalls}. Such methods usually involve \emph{training with a set of corruptions} to generalize to unseen ones~\cite{rusak2020simple}. However, performance gains can be non-uniform, e.g. Gaussian noise augmentation improves performance on other noise corruptions (e.g. impulse, shot noise) but hurts performance on fog and contrast~\cite{ford2019adversarial}. Instead, our main mechanism uses a large set of middle domains (not corruptions) to be resistant to a wide range of diverse \textit{unseen} corruptions. We do not use any corruptions during training, except to calibrate uncertainty.

\textbf{Adversarial attacks} add imperceptible worst case shifts to the input to fool a model~\cite{szegedy2013intriguing,kurakin2016adversarial,madry2017towards}. In contrast to \cite{pang2019improving,kariyappa2019improving,yang2020dverge}, which are ensemble based adversarial robustness methods with an additional loss term to promote diversity, the diversity of our ensembles is a natural consequence of using different middle domains. {\color{black} While our focus is not limited to robustness against adversarial attacks, it yields supportive evaluations against them as well (Sec. \ref{sec:results_adversarial}). 

}
\vspace{-0.0in}

\begin{figure*}[!ht]
\centering
  \includegraphics[scale=0.17]{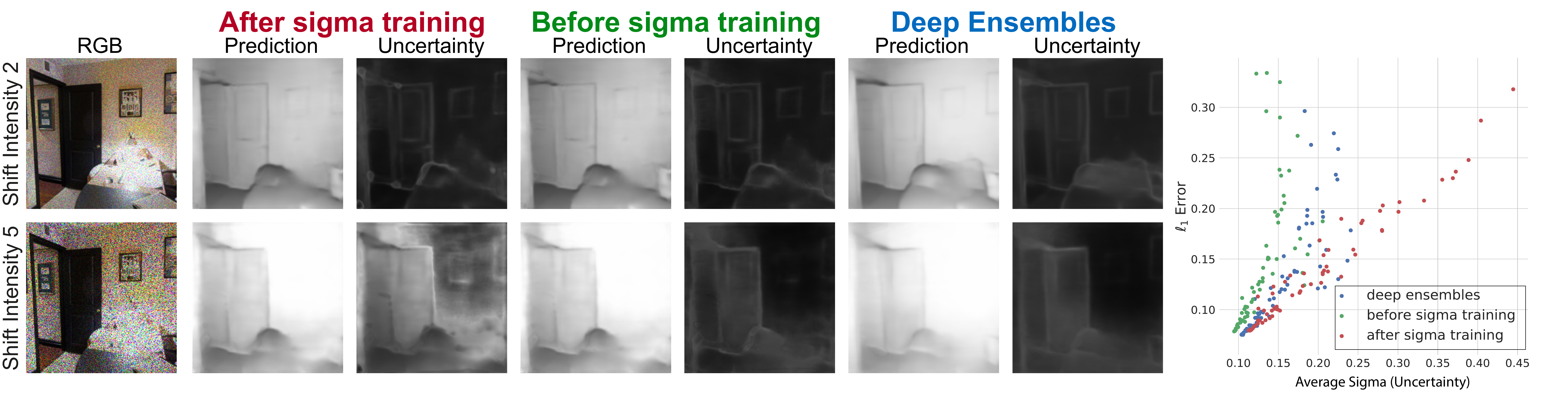}\vspace{-15pt}
\caption{\footnotesize{\textbf{Addressing overconfident inaccurate predictions under high distortions.} \textbf{Left}: Qualitative prediction results of image (re)shading and their corresponding uncertainty estimates (i.e. sigma) under two intensities of speckle noise distortion. This is shown for a single UNet model before and after \emph{sigma training} (ST) as well as for deep ensembles (standard deviation of predictions in the ensemble). Darker denotes lower uncertainty/sigma. ST was done using Gaussian noise and Gaussian blur distortions. \textit{Using other distortions yields similar performance}~(see \href{https://crossdomain-ensembles.epfl.ch/XDE_supp.pdf}{supplementary}). \textbf{Right}: Scatter plot of $\ell_1$ error versus average sigma. Each point is computed from an average over 16k test images for one of the \emph{unseen} distortions and one of 5 levels of shift intensity. Notice (qualitatively and quantitatively) that when the models without ST produce poor results, their uncertainty does \emph{not} correspondingly increase. Our ST helps the model to have a stronger correlation between its uncertainty estimates and error when \emph{tested on unseen distortions}. This indicates that sigma \emph{after ST} can be an effective signal for merging multiple predictions. {\color{black} Note that the predicted mean (``Prediction") \textit{does not change} with ST.} }}\label{fig:sigma}
\end{figure*}

\section{Method}

We explain the technical details of our method below.

\textbf{Notations:} Define $\gX$ as the RGB domain, $\gY=\{\gY_j\}_{j=1}^K$ as the $K$ intermediate domains, $\gZ$ as the desired prediction domain. A single datapoint $n$ from these $K$ domains is denoted as $(x_n,y_{1,n},\hdots,y_{j,n},\hdots,y_{K,n},z_n)$. $\mathcal{F}_\mathcal{XY}$ is the set of functions that maps the RGB images to their intermediate domains, $\gF_{\gX\gY}=\{f_j:\gX\rightarrow\gY_j\}_{j=1}^K$, and $\gF_\mathcal{YZ}$ is the set of functions mapping from the intermediate to the target prediction domain, $\gF_\mathcal{YZ}=\{g_{j}:\gY_j\rightarrow\gZ\}_{j=1}^K$. Given $K$ predictions of domain $\gZ$, they are merged using the function $m$ to get a final single prediction, $m:\{g_j(\gY_j)\}_{j=1}^K\rightarrow\gZ$.



\subsection{Estimating Per-Path Predictions and Uncertainty}\label{sec:method-noise}

We learn the mappings $g_j$ using a neural network. We model the noise in the predictions made by $g_j$ with a Laplace distribution. Thus, for an input sample $y_{j,n}$, the network outputs two sets of parameters $[\hat{z}_{j,n},\shat_{j,n}]=g_j(y_{j,n})$ where we set $\shat_{j,n}=\log \hat{b}_{j,n}$ for numerical stability and $\hat{b}_{j,n}$ is the scale parameter of the Laplace distribution. We remove the dependence on $j$ for brevity. This leads to the following negative log-likelihood~(NLL) loss for $g$:

\be
\mathcal{L}_{g,NLL} ~=~ \frac{1}{N}\sum_{n=1}^N \exp{(-\shat_n)} \Vert \hat{z}_n - z_n \Vert_1  + \shat_n \label{nll},
\ee
where $N$ is the number of samples, and $z_n$ is the label for the $n$th sample. This results in an $\ell_1$-norm loss on the errors as opposed to an $\ell_2$-norm loss with a Gaussian distribution, which has been shown to improve prediction quality~\cite{kendall2017uncertainties,zamir2020robust}. Finally, the \textit{sigma} is given by $\sqrt{2}\exp({\shat_n})$ and it captures the uncertainty in predictions.




\textbf{Calibration via Sigma training (ST):} Uncertainty estimates under distribution shifts are poorly calibrated~\cite{ovadia2019can}, i.e. there is a tendency to output a poor prediction with high confidence. This can be seen in Fig.~\ref{fig:sigma}, ``Before sigma training'' columns. With a higher noise distortion, the prediction clearly degraded, but the uncertainty estimate did not increase correspondingly. This issue persists even with methods that estimate epistemic uncertainty (Fig.~\ref{fig:sigma}, ``Deep Ensembles'' columns) which are meant to detect these shifts. 

To mitigate this, we adopt a two-stage training setup where the network trained on in-distribution data is further trained to output high uncertainty outside the training distribution. We denote this step as \textit{sigma training} (ST). Here, $\shat_n$ is trained to learn its maximum likelihood estimator, with the loss denoted as \textit{sigma calibration} (SC). As the goal of this step is to maximize the likelihood by correcting the sigma $\shat_n$, and not the mean $\zhat_n$, under a distortion (\textit{dist}), we add a loss term to ensure that $\zhat_n$ does not deviate from its predictions at the start of ST, which we define as $\zhat_{n}^0$. We denote this loss as \textit{mean grounding} (MG).  Finally, we include the original NLL from Eq.~\ref{nll} on undistorted data (\textit{undist}) to prevent forgetting. This results in the following loss formulation:
\be
\mathcal{L}_{g,ST} ~=~ \mathcal{L}_{g,NLL}^{undist} + \alpha_1\mathcal{L}_{g,MG}^{dist} + \alpha_2\mathcal{L}_{g,SC}^{dist},  \label{sig_aug}
\ee
where $\alpha_1, \alpha_2$ controls the weighting between the loss terms. For a given $\hat{z}_{n}^0$, the MG loss is defined as the $\ell_1$-norm distance between the current prediction and the one at the start of sigma training, i.e. $\mathcal{L}_{g,MG}^{dist} = \Vert \hat{z}_{n}^0 - \hat{z}_{n} \Vert_1$. The SC loss guides the scale parameter towards its maximum likelihood estimate, i.e. $\mathcal{L}_{g,SC}^{dist} =  \Vert \exp{(\shat_n)} - \argmin_{\shat_n} \mathcal{L}_{g_j,NLL}^{dist}  \Vert_1 = \Vert \exp{(\shat_n)} - |z_n - \hat{z}_{n}^0| \Vert_1$.

\begin{figure*}[!ht]
\centering
  \includegraphics[scale=0.105]{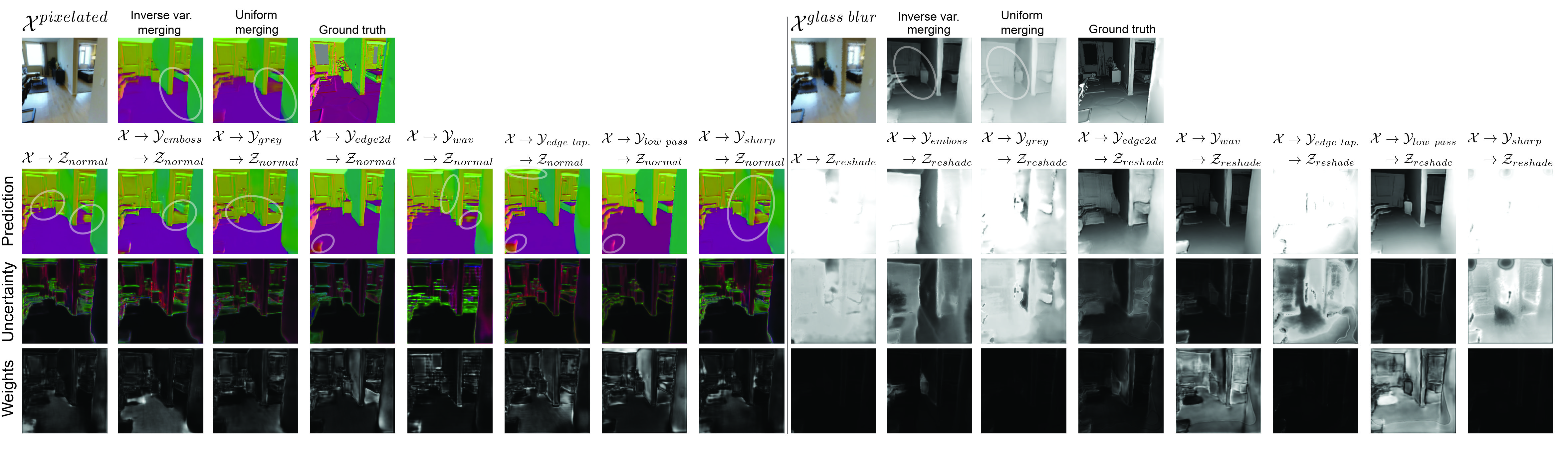} \vspace{-17pt}
\caption{\footnotesize{\textbf{How does the method work?} {Each network in each path receives different cues for making a prediction, due to going through different middle domains. \textbf{Left}: Given a distorted pixelated query, each path (columns) is affected differently by the distortion, which is reflected in its prediction, uncertainty, and weights (lighter means higher weights/uncertainty). The inverse variance merging uses the weights to assemble a final prediction that is better than each of the individual predictions. (The uncertainties of surface normals look colorful as surface normals domain includes 3 channels, thus there are 3 uncertainty channels.) \textbf{Right}: Similarly, for a query with glass blur distortion, the method successfully disregards the degraded predictions and assembles an accurate final prediction. Note that the proposed method (\emph{inverse var. merging}) obtains significantly better results than learning from the RGB directly (leftmost column of each example) which is the most common approach. The quality of the final prediction depends on the following elements: \textbf{1.} For each pixel, at least one middle domain is robust against the encountered distortion, and \textbf{2.} The uncertainty estimates are well correlated with error, allowing the merger to select regions from the best performing paths. \emph{Uniform merging} does not take into account the uncertainties and consequently lead to worse predictions. The elliptical markers denote sample regions where the merged result is better than all individual predictions.}}}\label{fig:how} \vspace{-10pt}
\end{figure*}

Following ST, the network outputs sigmas that are highly correlated with error (Fig.~\ref{fig:sigma}, rightmost plot). Given multiple predictions of the same target domain and their sigma estimates, this allows us to use the latter as a signal for merging to get a single strong prediction~(Sec.~\ref{sec:method-merge}).

As the objective of ST is to expose the network to inputs with high distortions as opposed to updating the final predicted mean, \textit{any} corruption with high intensity will suffice{\color{black}. The distortions used for ST are not the same distortions as the ones at test time. Please see \href{https://crossdomain-ensembles.epfl.ch/XDE_supp.pdf}{supplementary} Section~2.5 for a detailed study. Furthermore, the experiments~(Fig.~\ref{fig:sigma}, Fig.~\ref{fig:distortions_l1}, Table~\ref{table:adv}) indicate that sigma clearly generalizes to unseen distortions.

\subsection{Merging Predictions} \label{sec:method-merge}

After obtaining the set of mappings $\gF_{\gX\gY}$ and $\gF_\mathcal{YZ}$ with the method described above, it remains to merge the predictions coming from multiple paths using a merging function $m$. We employ an analytical approach given by $m(\{g_j(y_{j,n})\}_{j=1}^K)=C\sum_{j=1}^K \exp(-2\hat{s}_{j,n})\hat{z}_{j,n}$ where $C$ is a normalizing constant defined as $C=(\sum_{j=1}^K \exp(-2\hat{s}_{j,n}))^{-1}$. This performs a straightforward weighting of each pixel in each path by the inverse of its variance~\cite{hartung2011statistical} which can be done with negligible computational cost. 
We denoted this as \textit{Inverse variance merging} and will show in Section~\ref{sec:results_pixelwise} that it performs better than other analytical and learning based variants of our method.


The algorithm~\ref{algo:algo} summarizes our training procedure.


\begin{figure*}[!ht]
\centering
  \includegraphics[scale=0.145]{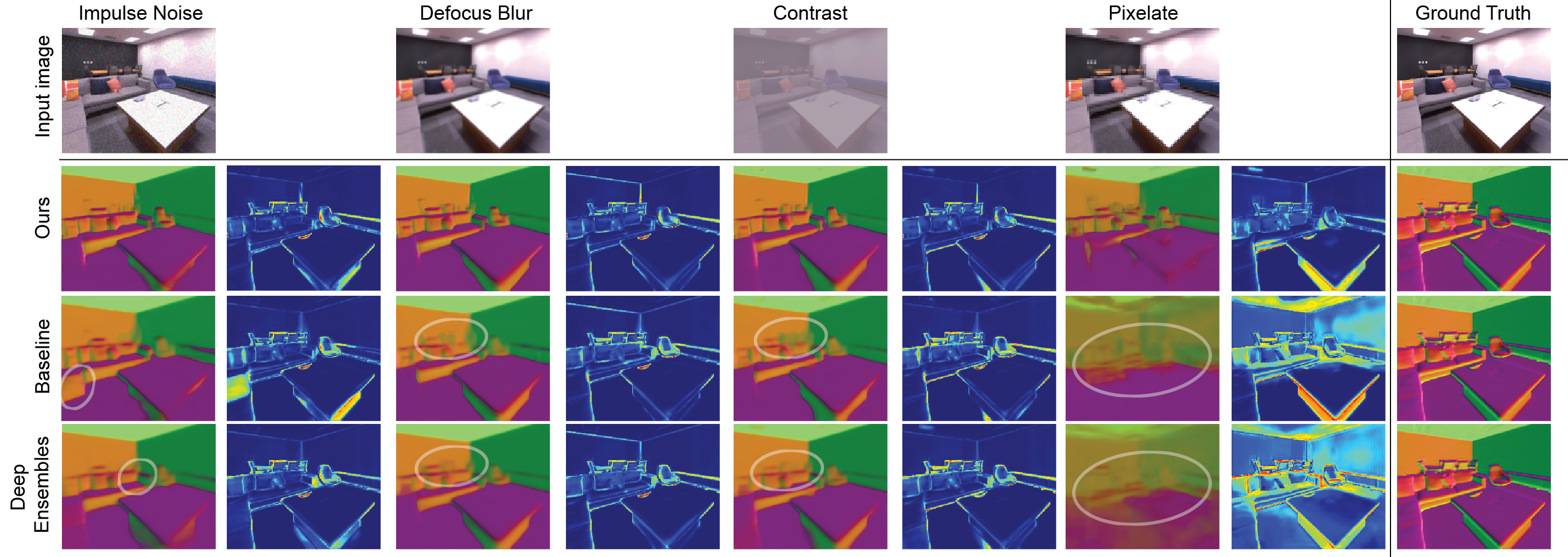} \vspace{-07pt}
\caption{\footnotesize{\textbf{Qualitative prediction results under 4 distribution shifts from Common Corruptions~\cite{hendrycks2019benchmarking}} shown on a sample image from the Replica\cite{replica19arxiv} dataset with shift intensity 3. {\color{black}Each prediction is followed by its corresponding error map.} Our method is resistant to distortions compared to the baselines and provides better accuracy \textit{especially over fine-grained regions and sharpness} (see the white markers). Best seen on screen.}}\label{fig:comparison-diverse} \vspace{-5pt}
\end{figure*}






\begin{algorithm}[H]

\caption{\small Summary of the training procedure of our method} \label{algo:algo}
\footnotesize
\begin{algorithmic}[1]
\REQUIRE Define $f_j \in \gF_{\gX\gY}$ and $g_j \in \gF_{\gY\gZ}$ $\forall j$.
\FOR{$j=1:K$}
\STATE Train $g_{j}$ using NLL loss in Eq.~\ref{nll}.
\STATE (Optional) Train $g_{j}$ using consistency constraints~\cite{zamir2020robust}. (Sec.~\ref{sec:results_pixelwise})
\STATE Perform sigma training over $g_{j}$ with Eq.~\ref{sig_aug}.
\ENDFOR
\STATE Merge the $K$ predictions from the $\gF_{\gY\gZ}$ networks using \textit{Inv. var. merging}~(Sec.~\ref{sec:method-merge}).
\end{algorithmic}
\end{algorithm}

\textbf{A working example.} Figure~\ref{fig:how} illustrates our method with an example. 
For a given image, each path's prediction, uncertainty, and corresponding weights are shown. For the distorted (pixelated) query in the left, each path reacted differently to the distortion, and the final prediction is obtained by combining individual predictions based on their uncertainties. Similar observations can be made for the glass blurred image in the right, where the method learned weights in a way such that the degraded paths are not used in the final prediction. We also show the final prediction from a uniform average of each path. While it is better than simply using the direct path ($\gX\rightarrow\gZ_{normal}$ or $\gX\rightarrow\gZ_{reshade}$), using the uncertainty estimates as weights results in a notably more accurate prediction.

There are two key elements to the effectiveness of our method. \textbf{I.} With a diverse set of middle domains, it is more likely that one of them will be less affected by distortions and returns an accurate prediction. \textbf{II.} The error of the prediction correlates well with its corresponding uncertainty estimates, i.e. the uncertainty is low in the region of the image where the prediction is accurate. This allows us to use these uncertainty estimates as a signal to have a final prediction with parts of the image taken from different paths.

\section{Experiments} \label{sec-exp}

We demonstrate that the proposed approach leads to robustness against different \emph{distribution shifts}, over different \emph{datasets}, and different \emph{prediction tasks}. For pixel-wise prediction tasks, we train on the Taskonomy dataset~\cite{zamir2018taskonomy}. To evaluate the robustness under corruptions, we report performance under Common Corruptions~\cite{hendrycks2019benchmarking} and adversarial perturbations~\cite{szegedy2013intriguing,kurakin2016adversarial,madry2017towards}. To evaluate against dataset shifts, we report on Replica~\cite{replica19arxiv} and Habitat~\cite{savva2019habitat} datasets. For classification, we train on ImageNet~\cite{russakovsky2015imagenet}, CIFAR~\cite{krizhevsky2009learning}, and evaluate on ImageNet-C and CIFAR-C~\cite{hendrycks2019benchmarking}. Please see the \href{https://crossdomain-ensembles.epfl.ch/XDE_supp.pdf}{supplementary} and \href{https://crossdomain-ensembles.epfl.ch}{project page} for more extensive qualitative results.



\subsection{Evaluations on Pixel-Wise Prediction Tasks}
\label{sec:results_pixelwise}
\textbf{Training dataset:}  We use Taskonomy~\cite{zamir2018taskonomy} as our training dataset which includes 4 million real images of indoor scenes with multiple annotations for each image. We report results for \textit{surface normals}, \textit{depth (zbuffer)}, and \textit{reshading} prediction, as popular target domains.

\textbf{Middle Domains:} From the RGB images we extract \textit{2D edges}, \textit{Laplace edges}, \textit{greyscale}, \textit{embossed}, \textit{low-pass filtered}, \textit{sharpened}, and \textit{wavelet} images as the middle domains (detailed definitions can be found in the \href{https://crossdomain-ensembles.epfl.ch/XDE_supp.pdf}{supplementary}). These middle domains {\color{black} are commonly used for low-level image processing tasks with negligible computation cost~\cite{acharya2005image,opencv_library} and} do not need any supervision. 
{\color{black} The performance was not sensitive to the choice of middle domains as the method consistently outperforms baselines and improves with more middle domains~(Sec.~\ref{sec:results_pixelwise_additional}, Fig.~\ref{fig:distortions_l12}).}

\textbf{Evaluation datasets:} Our goal is to have test data that has a distribution shift from the training data to evaluate the robustness of our method. All the results are reported on the test set of the following datasets:
\vspace{-1.5mm}
\begin{itemize}[leftmargin=*,label={}]
\setlength\itemsep{0em}
\item \textit{Taskonomy with Common Corruptions}~\cite{hendrycks2019benchmarking}:  We apply the Common Corruptions on the test set of Taskonomy. They include all corruptions except outdoor corruptions (snow, frost, fog) and the ones that change the geometry of the scene (elastic transform, motion, and zoom blur). {\color{black} We exclude Gaussian noise and blur from evaluations as they were used for ST, to keep training and testing fully separate.} Visualizations of a subset of distortions are shown in Figure~\ref{fig:comparison-diverse} and for all severities in the \href{https://crossdomain-ensembles.epfl.ch/XDE_supp.pdf}{supplementary}.
\item \textit{Taskonomy with Adversarial corruptions}~\cite{szegedy2013intriguing,kurakin2016adversarial,madry2017towards}: We generate adversarial examples using Iterative-Fast Gradient Sign Method (I-FGSM)~\cite{kurakin2016adversarial}.
\item \textit{Other datasets:} Replica~\cite{replica19arxiv} consists of 1227 images from high quality 3D reconstructions of indoor scenes. Similar to Taskonomy, we also apply common corruptions on these images. {\color{black} Habitat~\cite{savva2019habitat} consists of 1116 images from mesh renderings with a substantial shift from Taskonomy. We test on both datasets without fine-tuning~(see \href{https://crossdomain-ensembles.epfl.ch/XDE_supp.pdf}{supplementary}). }
\end{itemize}

\textbf{Training details:} All networks {\color{black} for our method and baselines} use the same UNet backbone architecture~\cite{ronneberger2015u} and were trained with AMSGrad~\cite{reddi2019convergence}. We used a learning rate of $5\times 10^{-4}$, weight decay of $2\times 10^{-6}$, and batch size of 64. The upsampling blocks of all networks resize the activation maps using bilinear interpolation. 

We also augment the network training with ``cross-task consistency constraints" (X-TC)~\cite{zamir2020robust} for generally better results, but this is not a fundamental requirement (ablation results provided in Sec.~\ref{sec:results_pixelwise_additional}). We follow~\cite{zamir2020robust} and apply non-probabilistic perceptual losses on the predicted mean.

\begin{figure*}[t!]
\centering
  \includegraphics[scale=0.125]{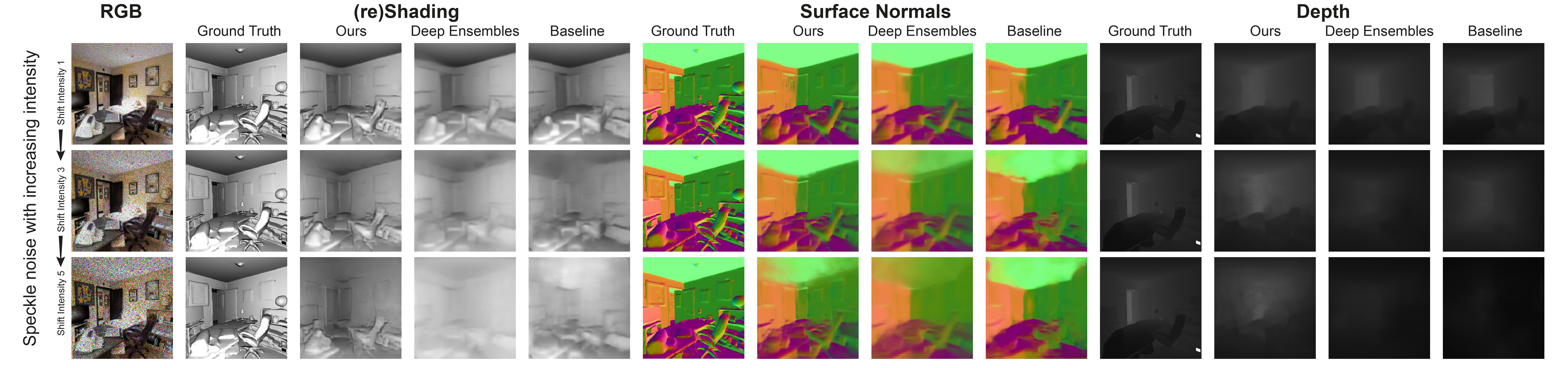}
  \vspace{-8mm}
\caption{\footnotesize{\textbf{Qualitative results under distribution shifts} for reshading, surface normals, and depth predictions. Each row shows the predictions from a query image from the Taskonomy test set under increasing speckle noise. Our method degrades less than the other baselines, demonstrating the effectiveness of using different cues to obtain a robust prediction. Notable improvements in the accuracy can be seen \emph{especially in fine-grained regions}. \vspace{-10pt}
}}\label{fig:comparison}
\end{figure*}

\textbf{Baselines:} We evaluate the following baselines. They are trained with NLL loss (Eq.~\ref{nll}), i.e. the models output both mean and sigma. 
\vspace{-2.mm}
\begin{itemize}[leftmargin=4.0mm,label={}]
\setlength\itemsep{-0.2em}
	\item \textit{Baseline UNet}: It is a single network that maps from RGB to the target domain without going through a middle domain~(i.e. direct). This is the main baseline. 
	\item \textit{Multi-domain baseline}: It is a network model with \textit{RGB} image \textit{and} all middle domains as inputs. Since this model is not \textit{forced} to use different middle domains as opposed to the proposed method, it reveals if learning from middle domains needs to be explicit and distributed.
    \item \textit{Multi-task baseline}: It is a single model that maps from \textit{RGB} to \textit{depth}, \textit{reshading}, and \textit{normals}. This is to reveal if learning additional tasks improved robustness.
    \item \textit{Data augmentation baselines}: We consider a baseline UNet adversarially trained to defend against I-FGSM attacks with $\epsilon=(0,16]$. This baseline shows how well adversarial robustness translates to non-adversarial distortions. We also include style augmentation~\cite{geirhos2018imagenet} as another baseline, which has been shown to reduce the texture bias that are less robust than shape cues.
	\item \textit{Blind guess} is a single prediction that captures the overall statistics of the domain, i.e. it returns the best guess of what the prediction should be independent of the input. Hence, it shows what can be learned from general dataset regularities (further details are in \href{https://crossdomain-ensembles.epfl.ch/XDE_supp.pdf}{supplementary}).
	\item \textit{Deep ensembles}~\cite{lakshminarayanan2017simple} creates an ensemble by training the same exact networks with different initializations. {\color{black} Although there are recent papers proposing new ways to enforce diversity in ensembles, their improvement in performance against deep ensembles has not been found significant under non-adversarial shifts~\cite{Wen2020BatchEnsemble,wenzel2020hyperparameter}. Thus, deep ensembles remains the most relevant ensemble baseline.}
	We use the same number of paths, i.e. ensemble components, as in our method. The predictions from each path are weighted equally to attain the final prediction. This baseline reveals if learning from different cues yields diverse predictions that results in a stronger final estimator.
\end{itemize}



\textbf{Cross-domain ensemble setups evaluated:} We evaluate several variants of our merging method. In all variants, different paths goes through different middle domain to produce a prediction along with one path being the \emph{direct} prediction. They are then merged into the final prediction. We show the proposed analytical merging is superior to others.
\vspace{-1.5mm}
\begin{itemize}[leftmargin=4.0mm,label={}]
\setlength\itemsep{-0.2em}
	\item \textit{Inverse variance merging}: Each path's prediction is weighted inversely proportional to its variance, as proposed in Section~\ref{sec:method-merge}. 
	\item \textit{Uniform merging}: A simplified merging where each path is weighted equally, i.e. uncertainty is not used.
	\item \textit{Network merging}: A neural network is used to merge the predictions. Specifically, we consider a stacking model~\cite{wolpert1992stacked} that learns the final predictions given the outputs from each path and models the final output as a mixture of Laplacians. It has the advantage that the loss is over the entire image, thus, taking into account its spatial structure~(see \href{https://crossdomain-ensembles.epfl.ch/XDE_supp.pdf}{supplementary} for details). 
	
\end{itemize}

\begin{figure*}[!ht]
\centering
  \includegraphics[width=\textwidth,keepaspectratio]{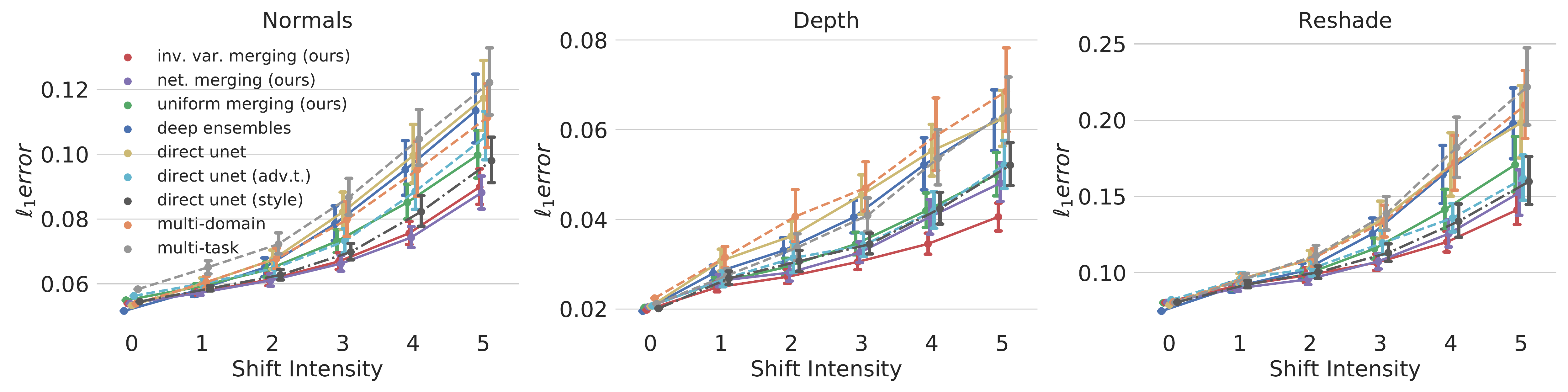} 
\caption{\footnotesize{\textbf{Quantitative robustness evaluations using Common Corruption distortions applied on Taskonomy test set: Average $\ell_1$ errors over 11 \emph{unseen} distortions.} Our main method \emph{inv. var. merging}, and frequently its simplified variant \emph{uniform merging} and \emph{network merging}, are more robust against shifts compared to the baselines. Error bars indicate one `standard error' from the mean (via bootstrapping). Plots for additional perceptual error metrics and individual distortions are provided in \href{https://crossdomain-ensembles.epfl.ch/XDE_supp.pdf}{supplementary material}. 
}}
\label{fig:distortions_l1}
\end{figure*}

\begin{figure}
\centering
\captionsetup{type=table} 
    \begin{adjustbox}{width=0.47\textwidth}
    \begin{tabular}{lllllllllllll}
     & \multicolumn{4}{c}{Normal} & \multicolumn{4}{c}{Reshade} & \multicolumn{4}{c}{Depth} \\ \cline{2-13} \hline
     \multicolumn{1}{|c|}{\diagbox{Method}{$\epsilon$}} & 2 & 4 & 8 & \multicolumn{1}{l|}{16} & 2 & 4 & 8 & \multicolumn{1}{l|}{16} & 2 & 4 & 8 & \multicolumn{1}{l|}{16} \\ \hline
     \multicolumn{1}{|c|}{Baseline UNet} & 8.23 &	11.53	& 13.03	 & \multicolumn{1}{l|}{14.37} & 17.92 &	22.78	& 27.26	&  \multicolumn{1}{l|}{34.40} & 5.50	& 6.76	& 8.36	&  \multicolumn{1}{l|}{9.80} \\
    \multicolumn{1}{|c|}{Deep ensembles} & \textbf{7.49} & 11.13 & 13.36 & \multicolumn{1}{l|}{15.65} & 15.66 & 21.95 & 27.75 & \multicolumn{1}{l|}{34.98} & 5.45 & 6.68 & 8.27 & \multicolumn{1}{l|}{10.52} \\
    \multicolumn{1}{|c|}{Inv. var. merging} & 7.60 & \textbf{8.89} & \textbf{10.40} & \multicolumn{1}{l|}{\textbf{12.77}} & \textbf{15.56} & \textbf{16.55} & \textbf{18.93} & \multicolumn{1}{l|}{\textbf{22.01}} & \textbf{4.94} & \textbf{4.99} & \textbf{5.93} & \multicolumn{1}{l|}{\textbf{6.75}} \\ \hline \hline
     \multicolumn{1}{|c|}{Adv. T. (lower bound error)} &  5.78 & 5.74	& 	5.45 & \multicolumn{1}{l|}{5.53} & 9.39 &	8.98 & 8.07	&  \multicolumn{1}{l|}{8.20} & 2.23 & 2.27	& 	2.39&  \multicolumn{1}{l|}{2.74} \\ \hline
    \end{tabular}
    \end{adjustbox} 
    \caption{\footnotesize{\textbf{Robustness against adversarial corruptions.}{~$\ell_1$ errors for surface normals, reshade, and depth under adversarial attacks are reported. (Lower is better. Errors are multiplied by 100 for readability.) The proposed method significantly improves robustness against I-FGSM~\cite{kurakin2016adversarial} based attacks \textit{without adversarial training}, compared to the baselines. The last row shows the error for a model that has undergone adversarial training~\cite{madry2017towards} with the \emph{same attacks as those evaluated at test time}, hence it gives a lower bound on the error~(see supplementary for additional details). \vspace{-10pt}}}}\label{table:adv}
    
\end{figure}


\subsubsection{Robustness to Common Corruptions} 
\label{sec:results_cc}
Figures~\ref{fig:comparison-diverse} and \ref{fig:comparison} show the qualitative results of our method against the baselines. Performance under various distortions is demonstrated in Figure~\ref{fig:comparison-diverse} for the surface normals predictions of a sample image from Replica dataset. The proposed method consistently outperforms the baselines and provides more accurate predictions especially in fine-grained regions. This is further supported by quantitative results in Figure~\ref{fig:distortions_l1} where the $\ell_1$ error over these distortions are notably lower for the proposed method compared to the baselines in all three target domains and shift intensities. 

{\color{black} Among the evaluated baselines the data augmentation methods are the most competitive, e.g. adversarial robustness partially transferred to non-adversarial distortions, though \textit{inverse variance merging} performs notably better. }

We also observe \textit{inverse variance merging} does much better than \textit{uniform merging} and also better or comparable to \textit{network merging}{\color{black}~(Fig.~\ref{fig:distortions_l1})} despite being simpler, more lightweight, and interpretable. Moreover, it does not demand fixing the number of paths beforehand (unlike \textit{network merging}), thus the number of paths can be decided by taking computational considerations into account on the fly.


\vspace{-6pt}

\subsubsection{Robustness to Adversarial Attacks} 
\label{sec:results_adversarial}
We demonstrate the effectiveness of the proposed method under adversarial attacks. The attacks are generated by I-FGSM. Following~\cite{kurakin2016adversarial}, we use attack strengths $\epsilon=[2,4,8,16]$, with the number of iterations given by $N=\min(4+\epsilon,1.25\epsilon)$. The results are shown in Table~\ref{table:adv}. Neither our method nor the baselines utilize explicit adversarial defense mechanisms -- while deep ensembles perform nearly as poorly as baseline UNet, the proposed method performs significantly better. This indicates that \textit{using middle domains} promotes ensemble diversity in a way that \textit{makes it more challenging to create one attack that fools all paths simultaneously}, hence this approach can be a promising remedy for adversarial attacks as well. Moreover, the proposed method also outperforms \textit{Uniform merging} (see \href{https://crossdomain-ensembles.epfl.ch/XDE_supp.pdf}{supplementary} for the results) which does not use uncertainty estimates during merging. This indicates that the additional uncertainty output did not create an additional avenue for attack that I-FGSM could exploit.

Note that we do not obfuscate gradients by e.g. intentionally making certain operations non-differentiable, or using stochastic transforms \cite{athalye2018obfuscated}. The analytical operations to obtain the middle domains are deterministic and differentiable. 




\begin{figure*}[th!]
 \begin{subfigure}[b]{0.60\textwidth}
    \centering
    \includegraphics[width=\textwidth,keepaspectratio]{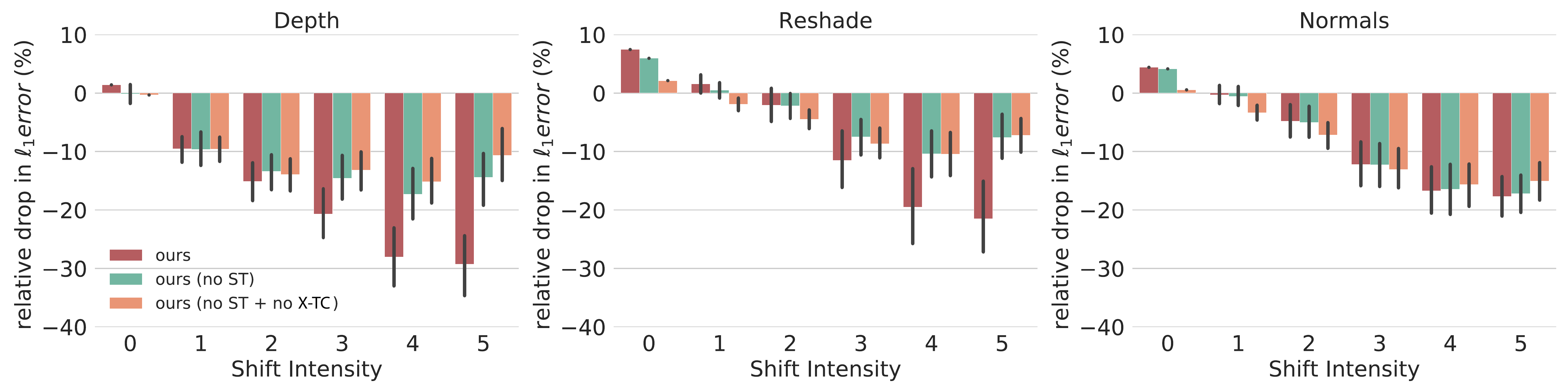}
    \includegraphics[width=\textwidth,keepaspectratio]{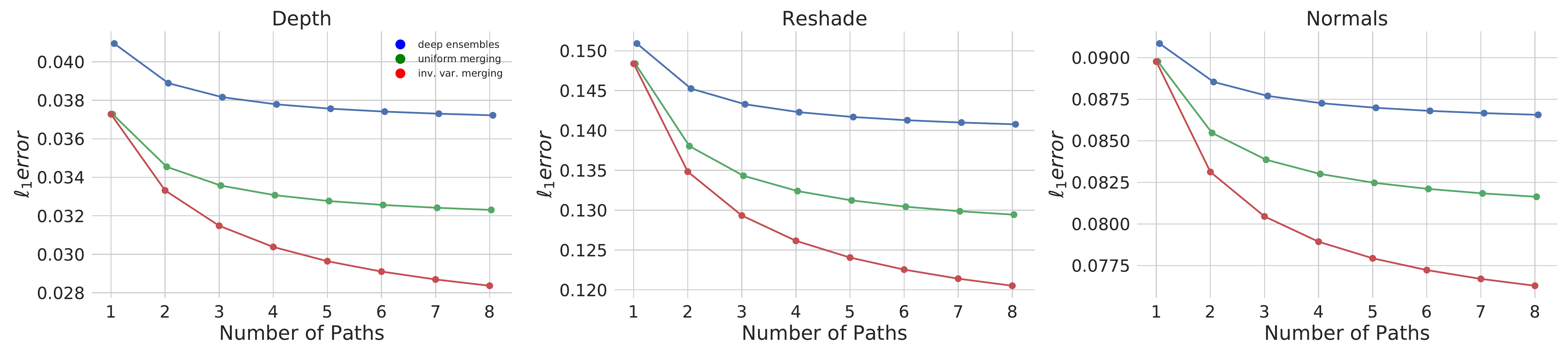}
    \caption{
    \footnotesize{
    \textbf{Effect of sigma and/or consistency training. (Top row)} The plots show the relative change in $\ell_1$ error compared to deep ensembles (i.e. negative means outperforming deep ensembles). The proposed method outperforms deep ensembles under distribution shifts even without ST and X-TC (consistency).
    \\
    \textbf{Robustness as a function of number of paths. (Bottom row)} The plots show the average $\ell_1$ error as we increase the number of paths (or ensemble components in the case of deep ensembles). The proposed method (\emph{inv. var. merging}) and its simplified variant (\emph{uniform merging}) consistently outperforms deep ensembles which plateaus much faster.
    }
    }\label{fig:distortions_l12}
 \end{subfigure}
 \hfill
 \begin{subfigure}[b]{0.37\textwidth}
    \centering
    
    \includegraphics[scale=0.33]{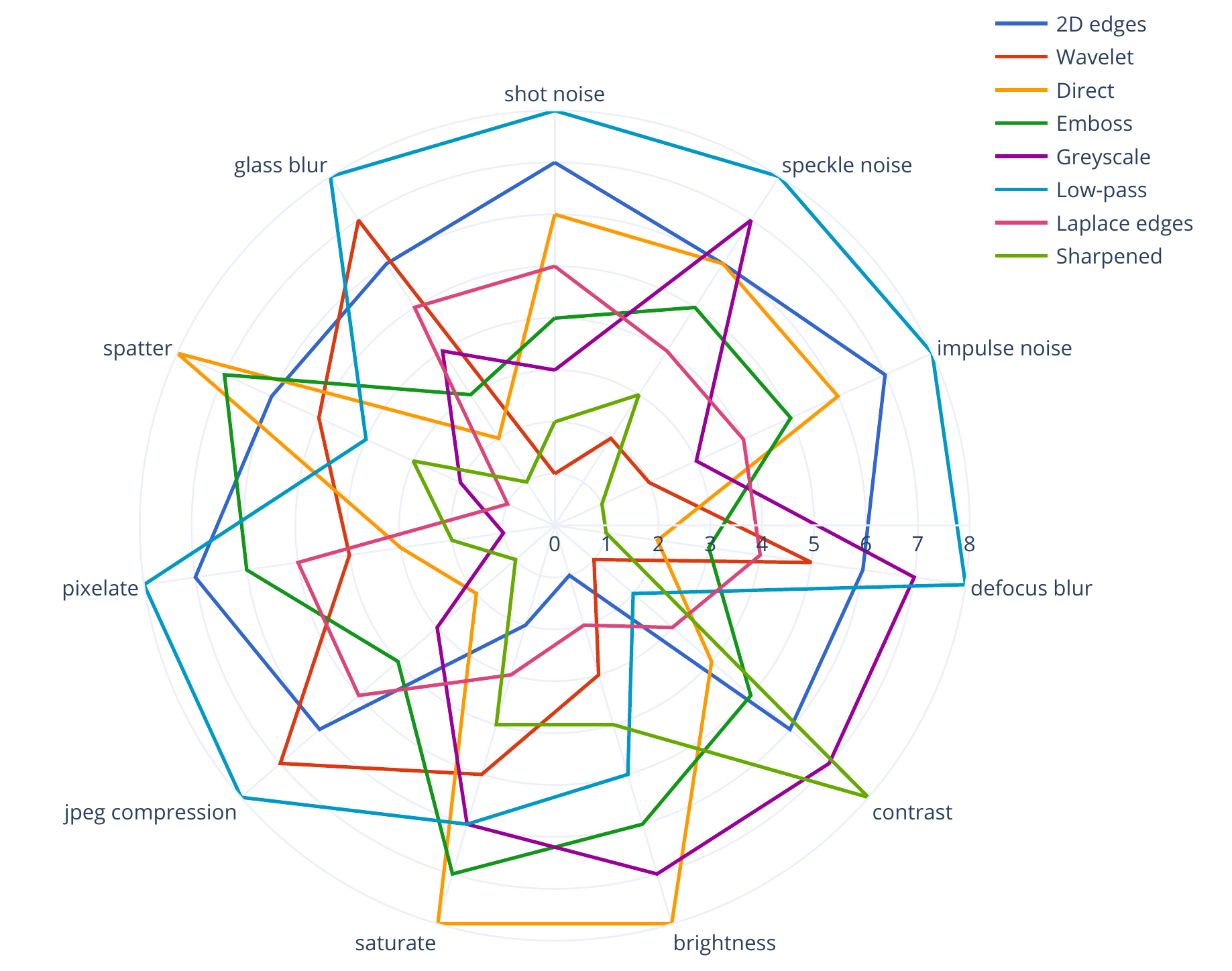}
    \caption{\footnotesize{\textbf{Importance of each middle domain for different distortions.} The chart shows the order of the best performing paths for surface normal perdiction for different distortions, with 8 denoting the most important and 1 the least important path. The plot shows, for instance, ``noise" distortions benefited most from the \textit{low-pass} middle domain while ``contrast" distortion benefited most from the \emph{sharpened} middle domain. 
    }}\label{fig:best_paths}
   \end{subfigure} \vspace{-5pt}
    
    \caption{\footnotesize{\textbf{Ablation studies:} We performed additional studies to gain insights on the effect of sigma and consistency training, increasing the number of paths, and the ordinal effect of each distortion on the middle domains. Similar to Figure~\ref{fig:distortions_l1}, in (a) and (b), we apply Common Corruption distortions on Taskonomy data and average the $\ell_1$ errors over 11 \emph{unseen} distortions. Error bars indicate one `standard error' from the mean (via bootstrapping). See \href{https://crossdomain-ensembles.epfl.ch/XDE_supp.pdf}{supplementary} for a more detailed breakdown of each of these studies. }}\label{fig:ablation}

\end{figure*}


\subsubsection{Additional Ablation Studies}\label{sec:results_pixelwise_additional}

\textbf{Contribution of ST/X-TC:}~~~To quantify the contribution of each stage of training to the overall robustness of our setup, we study the performance of our proposed method without sigma training (ST) and/or cross-task consistency constraints (X-TC) in the first row of Figure~\ref{fig:distortions_l12}~(and \href{https://crossdomain-ensembles.epfl.ch/XDE_supp.pdf}{supplementary}). 
Our method, with or without ST or X-TC constraints, still outperformed deep ensembles as almost all bars are below the 0 line. 

In Section 2.3 of the \href{https://crossdomain-ensembles.epfl.ch/XDE_supp.pdf}{supplementary}, we compare the effect of equipping deep ensembles with ST and X-TC, and perform uniform and inverse variance merging. Thus, the only difference with our method is the use of middle domains. Our method still outperforms.

\textbf{Robustness vs number of employed paths:}~~~In Figure~\ref{fig:distortions_l12}, we investigate performance as a function of number of paths. Each point shows the average $\ell_1$ error of \emph{all possible combinations} for a given number of paths. 
Although all methods improve as more paths are added, our proposed methods has a much steeper downward trend than deep ensembles and our uniform merging variant, indicating that performance gap increases with more paths.

\textbf{Sensitivity to choice of middle domains:}~~~Figure~\ref{fig:distortions_l12} also shows that the performance of our method is not sensitive to a particular set of middle domains. For a fixed number of paths $n$, our method outperforms deep ensembles for all possible combinations of $n$ paths on average.




\textbf{Path importance:} We show the importance of each middle domain for the final prediction under each distortion in Fig.~\ref{fig:best_paths}. For number of paths $n=1,\ldots,8$, we compute the set of best performing paths, i.e. the set of $n$ paths with the lowest $\ell_1$ error, denoted by $P_n=\{p_{(i)}\}_{i=1}^n$. The $n^{th}$ best performing path is given by $P_n\backslash P_{n-1}$. The plot shows different paths indeed react differently to a given corruption, e.g. noise distortions substantially benefited from \textit{low-pass}, while contrast distortion did not -- thus the benefit is not attributed to one or few middle domains under all distortions.

\subsection{Robustness of Sigmas to Distribution Shifts}\label{robustsig}

We have showed that our method returns predictions that are robust under a wide range of distribution shifts. Are our predicted uncertainties also able to generalize under distribution shifts, i.e. do we get high uncertainties when predictions get worse? To investigate this, we consider epistemic uncertainty which is used to capture the model's uncertainty and is an indicator of distribution shifts~\cite{kendall2017uncertainties}. The left plot of Figure~\ref{fig:sigma_merged} shows a scatter plot of average epistemic uncertainties against error and the right shows the average error for all epistemic sigma values less than a threshold $\tau$. The predicted uncertainties from deep ensembles initially increase with error but does not increase past 0.15 sigma despite an increase in error, thus is overconfident, while our method shows an increasing trend.

\begin{figure}[h!]
\centering \vspace{-5pt}
  \includegraphics[scale=0.2]{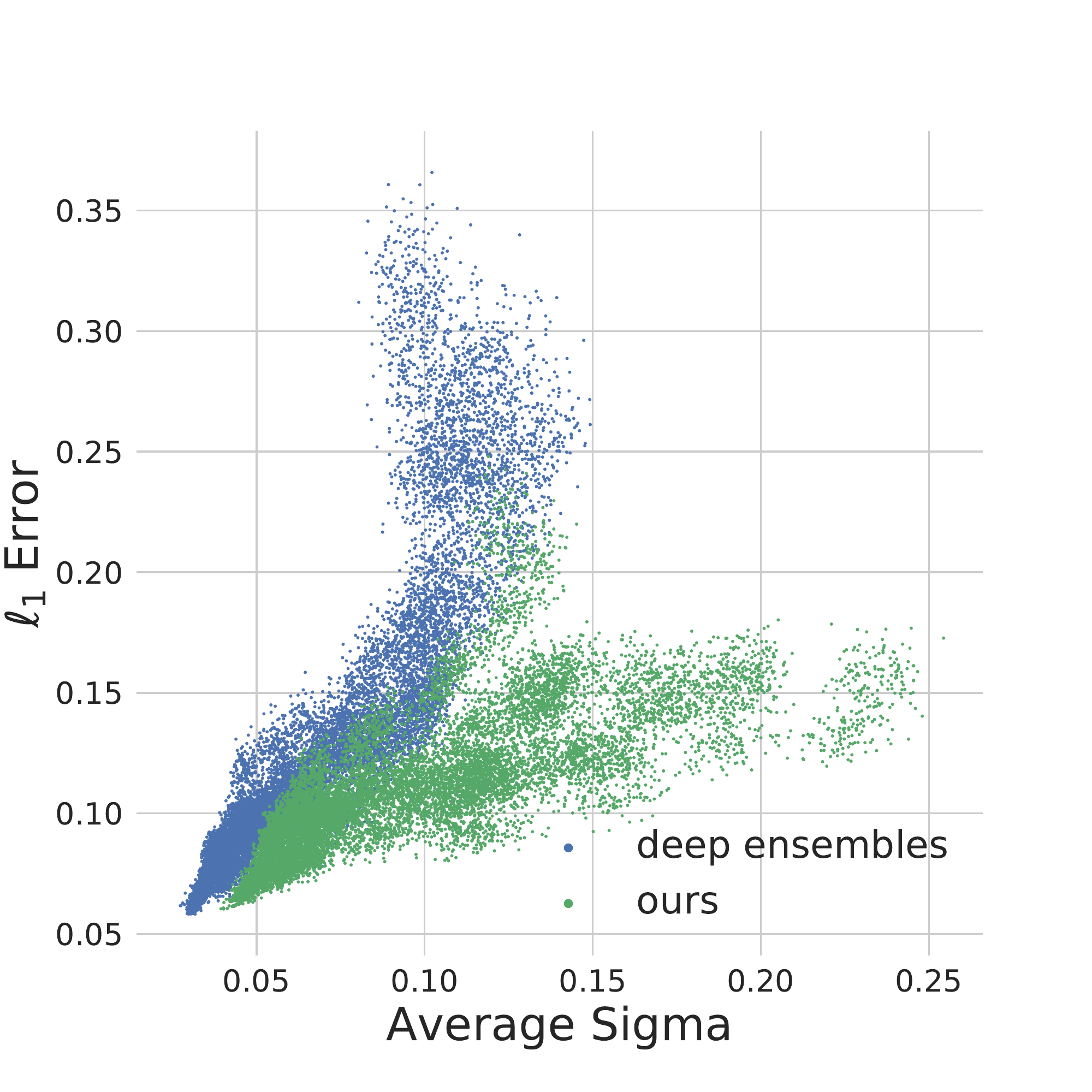} \hspace{-5pt}
  \includegraphics[scale=0.2]{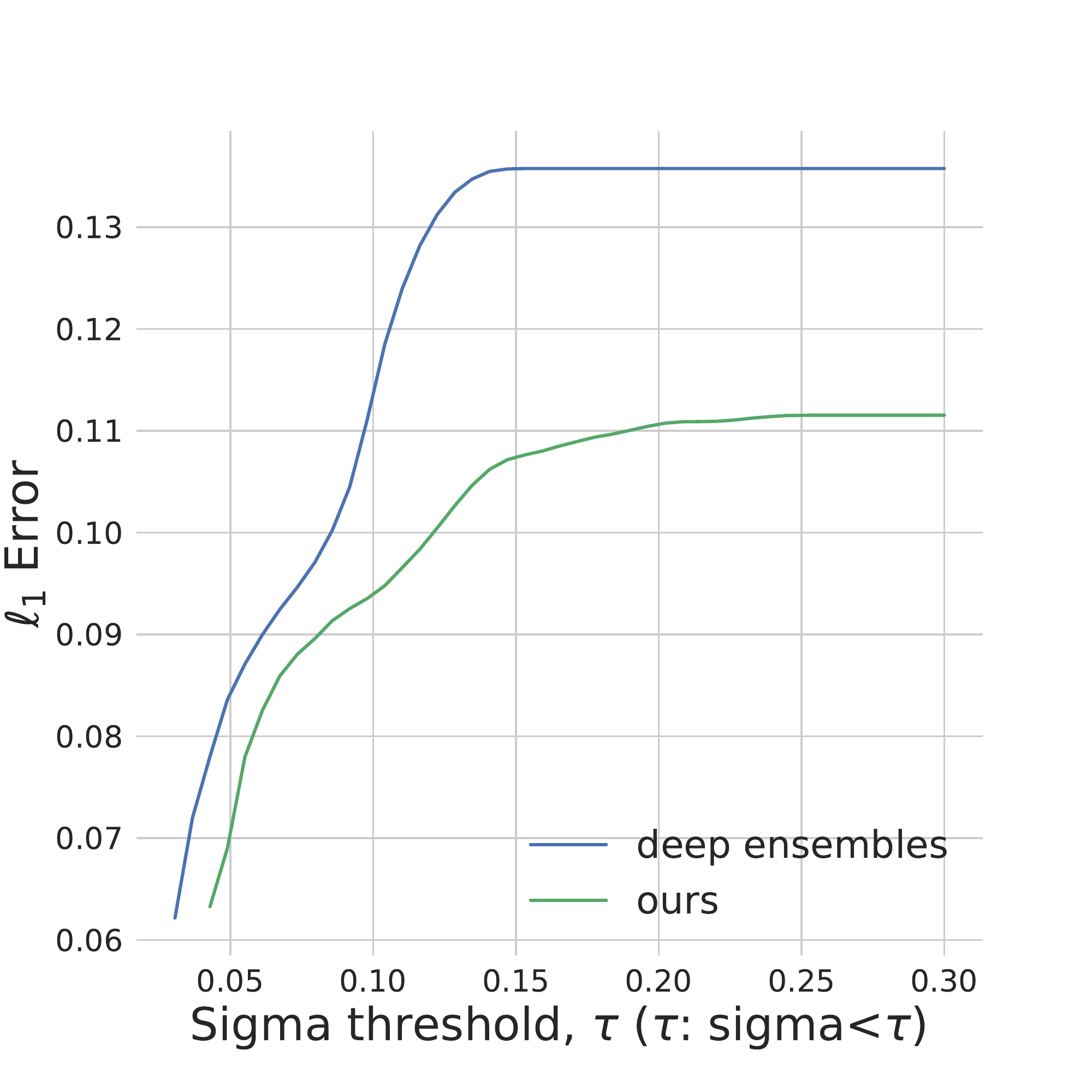} \vspace{-7pt}
\caption{\small{\textbf{Generalization of predicted uncertainties.} We compute the average $\ell_1$ error and epistemic uncertainties for 11 unseen distortions and 5 levels of intensities for Reshade. Each point is an average over 64 images. Our method is able to return \emph{high uncertainty when predictions are poor}.}}\label{fig:sigma_merged}
\end{figure}

\subsubsection{Performance on undistorted data}
In order to demonstrate that the robustness of our method on out-of-distribution data did not come at the cost of degraded performance on in-distribution data, we provide quantitative evaluations on the \textit{undistorted} Taskonomy and Replica datasets in \href{https://crossdomain-ensembles.epfl.ch/XDE_supp.pdf}{supplementary}. 
The results show the performance of our method, \emph{when tested on undistorted data}, is indeed comparable to or better than the methods that are trained to perform well only on undistorted data.


\subsection{Evaluation on Classification Tasks} 
\label{sec:results_classification}
The benefits of the proposed method is not limited to regression or dense pixel-wise tasks. We performed an experiment on ImageNet-C to evaluate the robustness against Common Corruptions (Table~\ref{table:imgnet-abs}). Our method and deep ensembles both use 8 paths with identical ResNet-50~\cite{he2016deep} network architecture. In addition, in this experiment our method does a simple averaging of the output probabilities from each path, similar to deep ensembles, and no ST or X-TC training was involved. The superior results show the basic value in using a diverse set of middle domains. {\color{black} Similar conclusions were obtained for CIFAR-10-C and CIFAR-100-C datasets~(full results in \href{https://crossdomain-ensembles.epfl.ch/XDE_supp.pdf}{supplementary}).}

\begin{table}[!ht]
\centering
\begin{adjustbox}{width=.25\textwidth}
\begin{tabular}{cl|l|} \hline
\multicolumn{1}{|c|}{Method} & \multicolumn{1}{l|}{Clean error} & \multicolumn{1}{l|}{mCE} \\ \hline
\multicolumn{1}{|c|}{Baseline ResNet-50} & 24.37 & 76.21 \\
\multicolumn{1}{|c|}{Deep ensembles} & {21.50} & 70.43  \\
\multicolumn{1}{|c|}{Ours} & 21.61 & {67.85}  \\ \hline
\end{tabular}
\end{adjustbox} \vspace{-7pt}
\caption{\footnotesize{\textbf{Robustness on ImageNet-C.}{~Error on clean and distorted data (mean Corruption Error -- mCE). Following~\cite{hendrycks2019benchmarking}, the mCE is relative to AlexNet~\cite{krizhevsky2012imagenet}. All methods are trained only on clean ImageNet training data. Our method performs noticeably better under distortions compared to deep ensembles and a single model baseline ResNet. {\color{black}See supplementary for a detailed breakdown and additional results on CIFAR.}
}}}\label{table:imgnet-abs}
\end{table}



\section{Conclusion and Discussion}

We presented a general framework for making robust predictions based on creating a diverse ensemble of various middle domains. Experiments demonstrated that this approach indeed leads to more robust predictions compared to several baselines. 

We also showed that our method is not sensitive to the choice of middle domains (Sec.~\ref{sec:results_pixelwise_additional}) or the corruptions used for ST (supplementary). Furthermore, even after equipping deep ensembles with ST and consistency training (Sec.~\ref{sec:results_pixelwise_additional}, supplementary), our method still outperforms, confirming the effectiveness of using middle domains.

Below we briefly discuss some of the limitations:  

\begin{itemize}[leftmargin=*,label={}]
    \setlength\itemsep{0em}
    \item \textit{Uncertainty under distribution shift}: Our method relies on having reasonable uncertainty estimates (i.e. sigma) in presence of distribution shifts. While we observed sigma training to be helpful for this purpose, and also, uniform merging which does not rely on uncertainty estimates to still outperform the baselines, our method will benefit from better uncertainty estimation techniques.
    \item \textit{Choice of middle domains}: We adopted a fixed set of middle domains, and, as discussed in Sec.~\ref{sec:results_pixelwise_additional}, the final performance was not sensitive to the adopted dictionary. However, \emph{learning} or computationally \emph{selecting} such middle domains with the objective of  downstream  robustness could be a worthwhile future direction.

    \item \textit{Multi-modal distributions}: We modeled our individual path outputs with single-modal distributions for convenience and considered multi-modal distributions only at merging step. Allowing for multi-modality in each path's output may further help with ambiguous data points.
   \item \textit{Computational cost:} While the computational complexity of our method and deep ensembles~\cite{lakshminarayanan2017simple} are virtually the same, the methods based on ensembling generally increase the computational complexity as they involve turning one estimator into multiple. Investigating if the models in the ensemble can be compressed would be worthwhile -- especially for our method since the diversity in the ensemble is by structure and owed to adopting different middle-domains, rather than stochasticities that often assume independence among models.
\end{itemize}

{\small
\bibliographystyle{ieee_fullname}
\bibliography{arxiv}
}

\end{document}